\documentclass[10pt,journal,compsoc]{IEEEtran}
\ifCLASSOPTIONcompsoc
  \usepackage[nocompress]{cite}
\else
  \usepackage{cite}
\fi
\ifCLASSINFOpdf
\else
\fi

\usepackage{ragged2e}
\usepackage{graphicx}
\usepackage{cite}
\usepackage{amsmath,amssymb,amsfonts}
\usepackage{algorithmic}
\usepackage{graphicx}
\usepackage{textcomp}
\usepackage{newtxmath}
\usepackage{booktabs}
\usepackage{graphicx}
\usepackage{float}
\usepackage{indentfirst}
\usepackage{listings}
\usepackage{subfig}

\begin{document}

\title{Metaknowledge Extraction Based on Multi-Modal Documents}

\author{Shukan Liu$^*$,
        Ruilin Xu$^*$,
        Boying Geng$^\dag$,
        Qiao Sun,
        Li Duan,
        and Yiming Liu

\thanks{$^*$ Contributed equally and should be considered as co-first authors.}
\thanks{$^\dag$ Corresponding author (e-mail: boying\_geng@126.com).}

\IEEEcompsocitemizethanks{\IEEEcompsocthanksitem Shukan liu was with the School of computer science and engineering, Southeast University, Nanjing, 211189, China; and the School of Electronic Engineering, Naval University of Engineering.\protect\\
E-mail: liusk@seu.edu.cn
\IEEEcompsocthanksitem Ruilin Xu and Yiming Liu were with the Graduate School, Naval University of Engineering, Wuhan, 430033, China; and the School of Electronic Engineering, Naval University of Engineering.\protect\\
E-mail: kirinxu@foxmail.com and liuyiming0507@foxmail.com
\IEEEcompsocthanksitem Boying Geng, Qiao Sun, and Li Duan were with the School of Electronic Engineering, Naval University of Engineering.}
}

\markboth{Preprint version on ArXiv}%
{Shukan Liu and Ruilin Xu \MakeLowercase{\textit{et al.}}: Metaknowledge Extraction Based on Multi-Modal Documents}

\IEEEtitleabstractindextext{%
\begin{abstract}
\justifying
The triple-based knowledge in large-scale knowledge bases is most likely lacking in structural logic and problematic of conducting knowledge hierarchy. In this paper, we introduce the concept of metaknowledge to knowledge engineering research for the purpose of structural knowledge construction. Therefore, the Metaknowledge Extraction Framework and Document Structure Tree model are presented to extract and organize metaknowledge elements (titles, authors, abstracts, sections, paragraphs, etc.), so that it is feasible to extract the structural knowledge from multi-modal documents. Experiment results have proved the effectiveness of metaknowledge elements extraction by our framework. Meanwhile, detailed examples are given to demonstrate what exactly metaknowledge is and how to generate it. At the end of this paper, we propose and analyze the task flow of metaknowledge applications and the associations between knowledge and metaknowledge. 
\end{abstract}

}

\maketitle

\IEEEdisplaynontitleabstractindextext

\IEEEpeerreviewmaketitle

\ifCLASSOPTIONcompsoc
\IEEEraisesectionheading{\section{Introduction}\label{sec:introduction}}
\else
\section{Introduction}
\label{sec:introduction}
\fi

\IEEEPARstart{R}{ecently}, widely used large-scale knowledge bases, such as DBpedia \cite{dbpedia}, FreeBase \cite{freebase}, and YAGO \cite{yago}, are all based on semantic entities and relations, focusing on specific concepts and things. These widely used large-scale knowledge bases have achieved good performance in information retrieval and question answering tasks. However, the very knowledge that exists in large-scale knowledge bases is based on entity-relation triples, which is not exactly as same as the knowledge in human beings' perception. Knowledge in human minds is a complex of hierarchical, structured, and systematized elements which has strongly logical or topological associations, especially presented in structure or sequence. 

For matching human’s natural intuition of knowledge, it is necessary to find a brand-new representation that makes knowledge in the computer structured and hierarchical. Thus, to address the significant issue that current knowledge bases have difficulty in forming knowledge hierarchy, this work introduces the concept of metaknowledge into knowledge engineering research. Ref. \cite{metaknowledge} studies metaknowledge in scientific articles and holds that "... metaknowledge results from the critical scrutiny of what is known, how, and by whom. It can now be obtained on large scales, enabled by a concurrent informatics revolution." Inspired by their thoughts, we believe that metaknowledge is knowledge about knowledge. It describes the patterns of how human beings get, analyze and organize knowledge. Similar to the metadata, metaknowledge is the structural representation of knowledge and knowledge with fine-grained and hierarchical characteristics.

Analogous to scientific articles in Ref. \cite{metaknowledge}, documents such as textbooks, papers, theses, governmental documents, laws, and regulations are the most commonly used materials in getting knowledge. Those materials are normally formed based on the author's logic and intentions which people can easily understand the backgrounds, motivations, themes only from structural patterns such as titles, section titles, keywords, etc. They are good data sources to acquire metaknowledge. Nevertheless, these structural patterns about getting knowledge cannot be well described by triple-based knowledge. This work holds that the structural elements such as titles, section titles, abstract, keywords and so on, potentially reflect the structural logical associations between knowledge. Therefore, in our study, the elements above are defined as Metaknowledge Elements. 

The process of people reading can be divided into three steps: see the words, know the words, and understand the words. When the computer imitates human beings’ process, the three steps become object detection, optical character recognition (OCR), and natural language processing (NLP). Object detection makes the computer classify and locate the layout elements of documents. OCR makes the computer know the exact textual content. NLP makes the computer analyze and understand the textual content. Documents of three modalities are involved in these signs of process: image modal, layout modal, and text modal. Therefore, a multi-modal method should be designed to extract metaknowledge elements from documents. 

To extract metaknowledge elements from multi-modal documents, our work proposes MEF, a multi-modal \textbf{M}etaknowledge \textbf{E}xtraction \textbf{F}ramework. Through experiments on multi-modal Chinese governmental documents dataset GovDoc-CN, MEF performs better than single-modal models. Furthermore, this work refines and explains the concept, the functions, the applications task flow of metaknowledge, and proposes that the metaknowledge applications are in a higher level compared with knowledge applications. Metaknowledge applications are similar to the knowledge applications partly but play a more significant role in managing and organizing knowledge.

\section{Related Works}

\textbf{Visually-Rich Document Analysis (VRDA)} is a task aiming to analyzing visually-rich documents (VRDs) which are scanned or digital-born, such as page images or PDFs, it plays a crucial role in governmental and commercial applications. VRDA basically relies on three modalities: the text modal, the layout modal, and the image modal. Ref. \cite{vrdcnn} presents an end-to-end and multi-modal fully convolutional networks to extract document semantic elements. Ref. \cite{vrdgcn} introduces a graph convolutional model that combines textual and visual information in VRDs. Ref. \cite{vrdpretrain} combines large pre-trained language models and graph neural networks to encode both textual and visual information in VRDs. Ref. \cite{vrdwebpage} presents an approach based on graph neural network to build a rich representation of text fields on a webpage and the relationships between them. The works above have shown significant improvement on VRDA, whereas they do not comprehensively consider the combination of all three modalities (text, layout, and image). LayoutLM \cite{layoutlm1} and its improved model \cite{layoutlm2} are the pre-trained and fine-tuned multi-modal framework to analyze VRDs, it unites text modal, layout modal, and image modal to better extract semantic elements. Nonetheless, the textual information in LayoutLMs is extracted from the image region it corresponds to by OCR, that is, the frameworks consider textual information of a document as independent semantic units instead of a coherent passage. 

\textbf{Hierarchical Graph} is initially proposed in the document reading comprehension. Focusing on natural questions answering, Ref. \cite{mhhg} presents multi-grained machine reading comprehension framework for modeling documents with their hierarchical natures. The framework in Ref. \cite{mhhg} divides a document into four levels of granularity: document, paragraphs, sentences, and tokens, then utilizes graph attention networks (GATs) to obtain a multi-grained representation of the document. Meanwhile, the recent work \cite{hgn} presents hierarchical graph network (HGN) for the multi-hop question answering task. Different from the one-hop question answering that answers originated from a single paragraph, the multi-hop question answering focuses on acquire answers from multiple paragraphs or the whole passage. HGN is built by constructing nodes on different levels of granularity including questions, paragraphs, sentences and entities, it aggregates clues from texts across multiple paragraphs. In creating HGN, the initial node representations are updated through graph propagation and traverses through the graph edges of subsequent sub-tasks for multi-hop reasoning. The works above present exquisite models for modeling documents, but restricted in a single modal of text. 

To make the computer effectively organizes and utilizes documents described by natural language, the first task is to convert nonstructural documents in natural language into structural data, where \textbf{Natural Language to SQL (NL-to-SQL)} is one of such the task. Approaches (Seq2SQL \cite{seq2sql}, SQLNet \cite{sqlnet}, RE-SQL \cite{resql}, etc.) and datasets (SPLASH \cite{splash}, ACL-SQL \cite{aclsql}, etc.) all aim to transform natural language to a structural data format, which generally is the Structured Query Language (SQL). The natural language materials these approaches and datasets processed are commonly tokens and sentences, while whereby the passage-level information is not referred to. Recently, Ref. \cite{doc2graph} designs a weakly-supervised text-to-graph model \textbf{Doc2Graph} to bridge the gap between concept map construction and neural networks, it is able to translate passage-level documents into graph data. 

Thanks to the inspirational works above, in this work, we consider getting through the key links between how to extract the semantic structure of documents (where VRDA focuses on), how to model structured documents (where HGN focuses on), and how to translated documents in natural language into data that the computer can understand (where NL2SQL and Doc2Graph focus on). 

\section{Metaknowledge Extraction Framework}
In this work, a multi-modal framework is designed to extract metaknowledge elements from documents in the text modal and the image modal, which is described as Fig. 1. There are three parts in the framework: (1) metaknowledge elements extraction modules, which extract metaknowledge elements form documents in both text modal and image modal, such as titles, subtitles, authors’ information, etc.; (2) verification and alignment module, which aligns metaknowledge elements from text modal and image modal; (3) metaknowledge generating module, which organizes metaknowledge elements by document topology and hierarchy, then generates metaknowledge through entity recognition and relation extraction.

\begin{figure*}[t!]
\centering
\includegraphics[width=\linewidth]{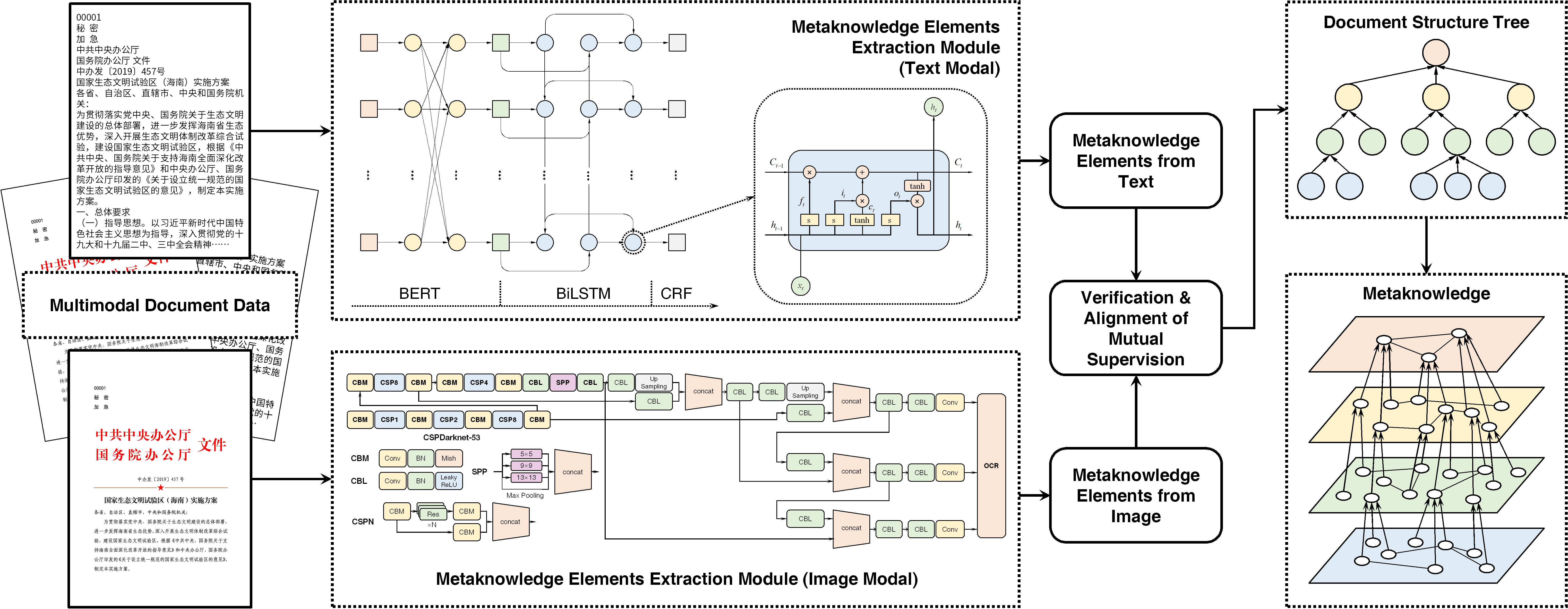}
\caption{The metaknowledge extraction framework (MEF), including: (1) Metaknowledge elements extraction modules (from both text modal and image modal); (2) Verification and alignment module; (3) Metaknowledge generating module. The text modal extraction uses BERT+BiLSTM+CRF \cite{bert_bilstm_crf} framework, the image modal extraction uses YOLOv4 \cite{yolo4} + PaddleOCR \cite{ppocr} framework.}
\vspace{-0.3cm}
\label{fig_1}
\end{figure*}

\subsection{Text Modal Metaknowledge Elements Extraction}
BERT \cite{bert} is a pre-trained transformer model. It designs two pre-training tasks: Masked Language Model (MLM) and Next Sentence Prediction (NSP). In MEF, BERT is applied to vectorize the text in the text modal. When text modal document $\boldsymbol{D}_t$ is input, BERT transform it into vector $\boldsymbol{v}_t$.
 
The BiLSTM structure adds a layer of reverse LSTM network based on a unidirectional LSTM network to use the context information more effectively. The BiLSTM network can better capture the preceding and subsequent messages at a certain time. It calculates two different hidden layer representations for each sentence using the sequential and reverse order local methods and then obtains the final hidden layer representation by splicing vectors.

In the task of NER, BiLSTM can predict the probability of each input word corresponding to different output labels. For these directly obtained probabilities, we can judge the label with the highest probability of the current token. However, only using the BiLSTM network will make the output ignore the correlation between labels and only pay attention to the association between input characters and labels, which will lead to great deviation in recognition results. To solve the problem, a conditional random field (CRF) is added in BiLSTM. The BiLSTM+CRF model \cite{bilstmcrf} uses the CRF layer to add constraints, which effectively reduces the false prediction results of BiLSTM. 

\vspace{-0.2cm}
\begin{figure}[!h]
\centering
\includegraphics[width=\linewidth]{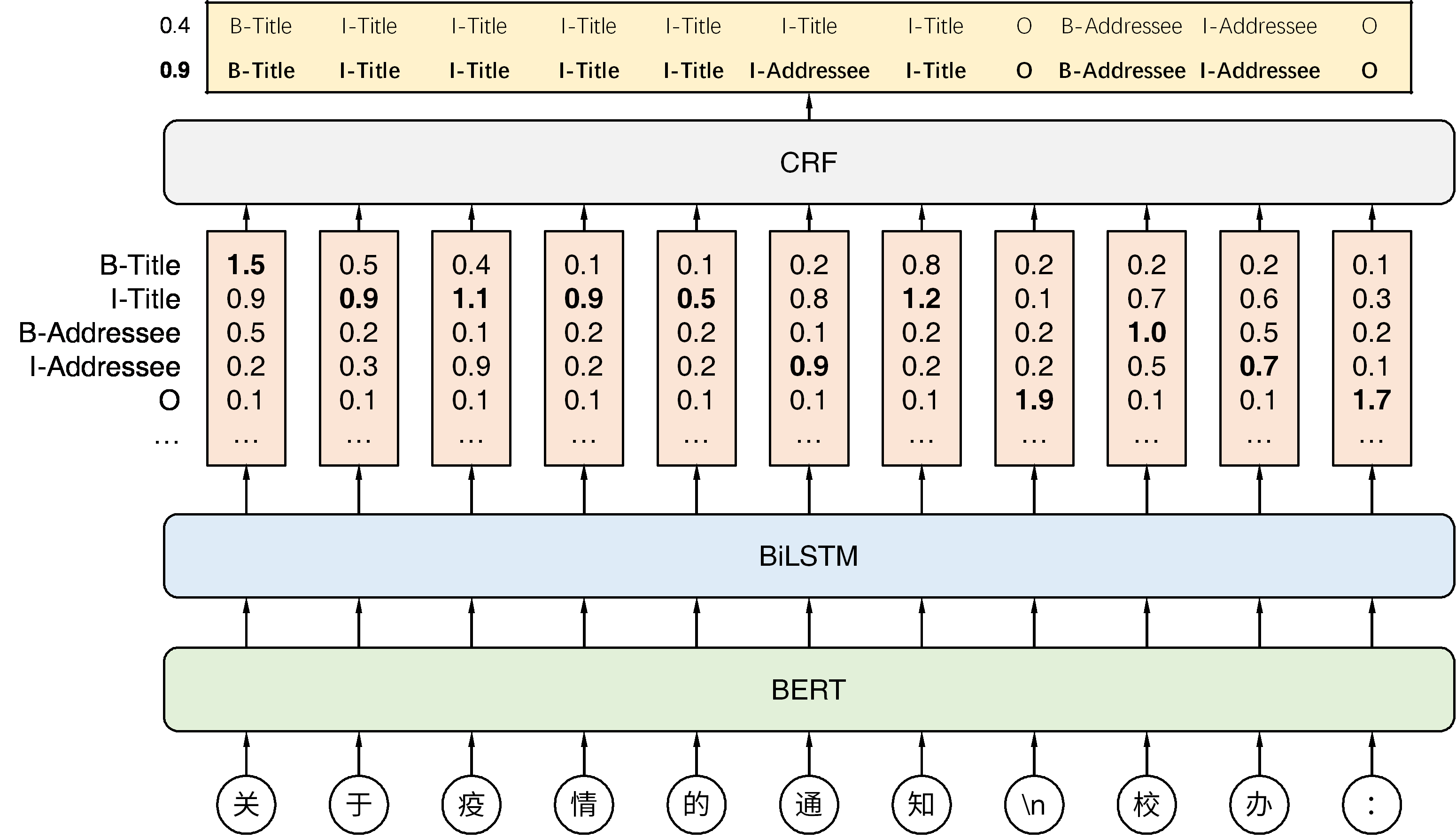}
\caption{An example of BERT+BiLSTM+CRF model: The input Chinese text sequence means “Notice on epidemic situation (line break) University Office”, in which “Notice on epidemic situation” is the title and “University Office” is the addressee. All the tokens in the sequence are annotated in BIO format.}
\vspace{-0.2cm}
\label{fig_2}
\end{figure}

Inspired by the above works, in this paper, we use BERT+BiLSTM+CRF \cite{bert_bilstm_crf} (Fig. 2) to extract metaknowledge elements from text modal documents in a way similar to NER. The word vector output from the BERT is input into the BiLSTM network. The output of BiLSTM is spliced and integrated, and the CRF layer is superimposed behind the BiLSTM to increase the utilization of label information. The CRF layer uses the transfer feature to consider the correlation between labels, combines BiLSTM and CRF to obtain the global optimal output sequence so that the prediction label is a sequence considering the label correlation. 

\subsection{Image Modal Metaknowledge Elements Extraction}
In the image modal, the metaknowledge elements extraction is similar to the document layout analysis task, which is a combination of optical character recognition (OCR) and object detection. Specifically, it is to find the regions of the metaknowledge elements through the object detection and then recognize the text content of the metaknowledge elements from regions they belong to. 

The image modal metaknowledge elements extraction module is designed based on the YOLOv4 \cite{yolo4} framework and PaddleOCR \cite{ppocr}. As is shown in Fig. 1, the extraction module includes an object detection part and an OCR part. The object detection part uses YOLOv4 framework to detect metaknowledge elements in document image. Each metaknowledge element corresponds to a vector $\boldsymbol{v}_i = (l_i, p_i, x_i, y_i, w_i, d_i)$, where $l_i$ represents the label of the object, $p_i$ indicates the probability that the object belongs to this type of element, $(x_i,y_i,w_i,d_i)$ is the bounding box coordinate of the location of the target, representing the location origin $(x_i, y_i)$, width $w_i$ and height $d_i$. 

Due to the accuracy of OCR, the text recognition result of the detected image region can not be guaranteed to be accurate, but can be corrected by using the corresponding text modal data. In this work, the Levenshtein distance (Eq. 1) between the OCR extraction results and the original texts in the text modal is calculated for correcting the OCR errors.

In the OCR post-process above, for each element of image recognition, the text element with the shortest Levenshtein distance corresponding to it will be found. Considering the possible errors in the optical character recognition module for the extraction of image modal information, this work holds that for general elements, when the Levenshtein distance between OCR recognition results and corresponding named entity recognition results is not more than 3 of OCR recognition results, that is, the difference is not more than 3 of the total sentence length of image recognition, the two are consistent. 

\begin{equation}{
lev_{a,b}(i,j) = \left\{ \begin{array}{l}
\max (i,j),if\min (i,j) = 0;\\
else \min \left\{ \begin{array}{l}
le{v_{a,b}}(i - 1,j) + 1\\
le{v_{a,b}}(i,j - 1) + 1\\
le{v_{a,b}}(i - 1,j - 1) + 1\;({a_i} \ne {b_j})
\end{array} \right.
\end{array} \right.
\label{eq1}
}
\end{equation}

\subsection{Multi-Modal Metaknowledge Elements Alignment}
The single-modal approaches have different performances when extracting various metaknowledge elements. Therefore, it is necessary to consider the two modalities' extraction results at the same time to fix the fault-tolerance problem of single-modal extraction to improve the extraction quality.

For metaknowledge element $i$, let the extraction result of text modal be the One-Hot vector $\boldsymbol{v}_{i\_text}$,  the extraction result of image modal be the One-Hot vector $\boldsymbol{v}_{i\_image}$. If the total number of metaknowledge elements is $n$ and the total number of element categories is $k$, then the total number of all possible combinations of the multi-modal extraction results is $k^2$. Assume the $k^2 \time 2$ matrix $\boldsymbol{W}$ is the decision matrix, $\boldsymbol{W}=[w_{j1},w_{j2}]_{k^2 \times 2}$. Each row in $\boldsymbol{W}$ represents a possible combination of multi-modal extraction results. Suppose the j-th row in $\boldsymbol{W}$ represents the case that "the extraction result of the element $i$ in the text modal is the $k_1$ type, and the extraction result of the same element $i $ in the image modal is the $k_2$ type", where $j=k_1 \times k_2$. If the extraction result of the text modal is correct but the extraction result of the image modal is incorrect, set $w_{j1} = 1, w_{j2} = 0$, otherwise, set $w_{j1} = 0, w_{j2} = 1$. If the extraction results of the two modalities are both correct, then set $w_{j1} = w_{j2} = 0.5$. Therefore, for the metaknowledge element $i$, the final extraction result is $\boldsymbol{v}_{i\_text}=w_{j1}\boldsymbol{v}_{i\_text}+w_{j2}\boldsymbol{v}_{i\_image}$. When $\boldsymbol{W}$ is trained with enough samples, the extraction results in the two modalities ($k_1$ and $k_2$) only need to look up the $k_1\times k_2$ row of matrix $\boldsymbol{W}$, and the weighted summation is the final extraction result.

\subsection{Organizing Metaknowledge Elements}
The framework proposed above realizes the classification of various types of metaknowledge elements. Still, it does not clarify the juxtaposition and inclusive relationships between the elements, especially the sections at different levels, and does not form a hierarchical document structure. 

From the perspective of people's writing and reading habits, to determine the relationship between sections at different levels, it is only necessary to consider the order in which sections appear in the whole document. Among the sections of different levels belonging to the "inclusive" relationship, the level of the section that appears first must be higher than that of the later; among the sections of the same level belonging to the "juxtaposition" relationship, the order in which they appear in the document can also reflect the relationship. Generally speaking, it is to rank sections with the "inclusive" relationship and to sequence sections with the "juxtaposition" relationship, through which different levels of sections appear in the document.

In MEF, through segmenting paragraphs and sentences, the document is divided into a set $T_{Document}=\{(x_0,y_0),(x_1,y_1),\cdots,(x_n,y_n)\}$, where the coordinate $(x,y)$ is the Paragraph Index (PI) and Sentence Index (SI) of each sentence in the document. For instance, assume the coordinate of sentence $i$ is $(x_i,y_i)$, where $x_i$ represents that sentence $i$ belongs to the $x_i$-th paragraph in the whole document, $y_i$ represents that the sentence $i$ is the $y_i$-th sentence in the $x_i$-th paragraph. The smaller the PI, the paragraph in which the sentence is in the front; the smaller the SI in the paragraph, the more forward the sentence in the paragraph.

Exact Subgraph Enumeration Tree \cite{esutree} (ESU-Tree) is a structural model designed for network motif recognition. This model is used to search for subgraphs of a specified scale in the network. Since the structure of ESU-Tree can better reflect the hierarchical and structural relationship, based on ESU-Tree, we propose a special structure for the hierarchical representation of documents, which is called "Document Structure Tree (DST)" (Fig. 3).

\begin{figure}[!h]
\centering
\includegraphics[width=\linewidth]{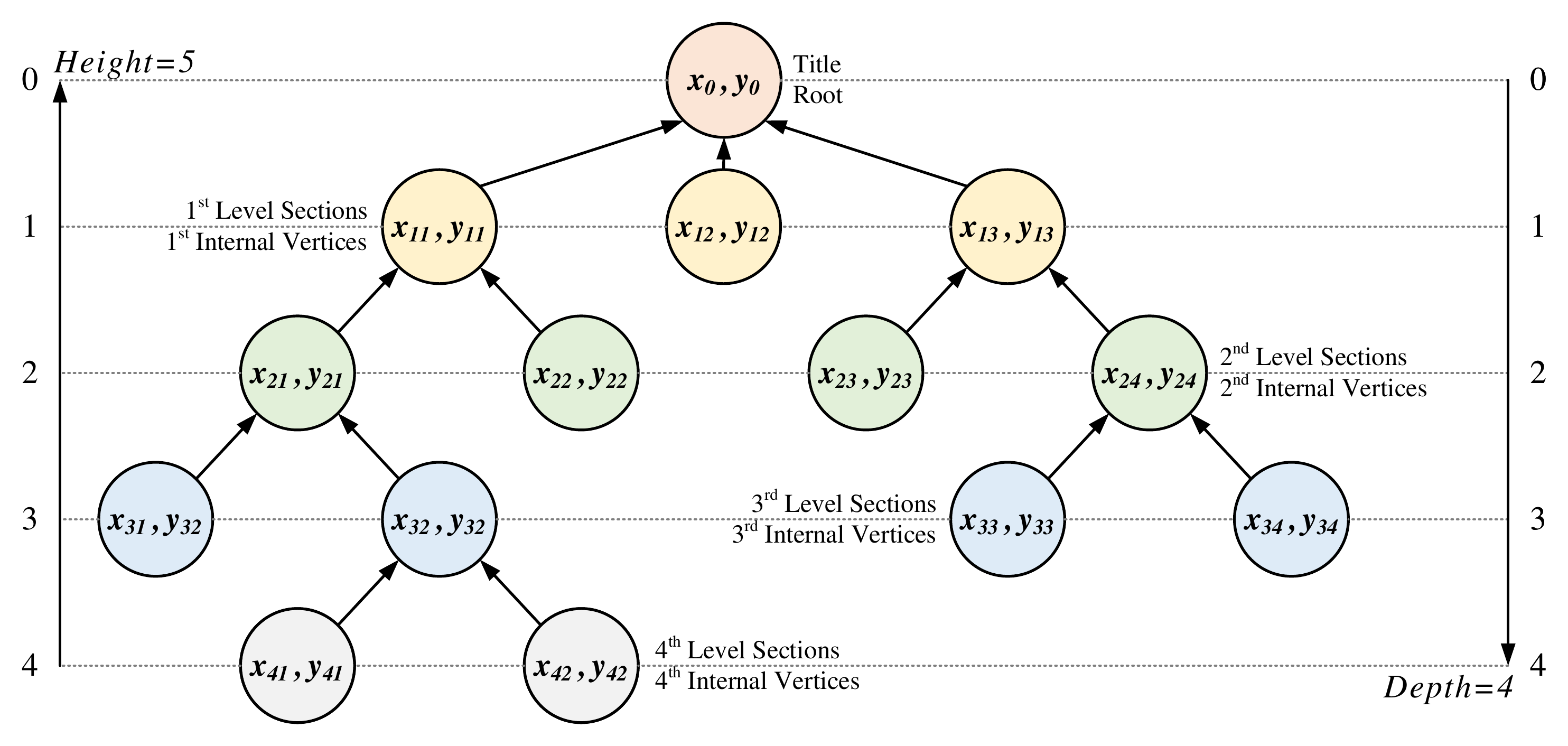}
\caption{Document Structure Tree}
\label{fig_3}
\end{figure}

\noindent\textbf{DEFINITION 1}\quad The Document Structure Tree (DST) is a directed rooted tree that:
\begin{itemize}
    \item Each of its child vertices points to its parent vertex; 
    \item Its root vertex is located at level 0, the number of full tree levels is 4, the depth is 4, and the height is 5; 
    \item Its fourth layer is all leaf vertices; 
    \item Its vertices have weights, but edges have no weights, and the vertex weights are composed of two parts \textbf{front weight} and \textbf{back weight}. When comparing weights, the front weights is compared first, and the back weights is compared when the two front weights is equal; 
    \item Its left vertex has a smaller weight than its right vertex, and its parent vertex has a smaller weight than its child vertices.
\end{itemize}

\noindent\textbf{DEFINITION 2}\quad If the vertex $RP$ is the nearest right vertex of a $Child$ vertex's parent, define $RP$ as the \textbf{Right Parent} of $Child$. Similarly, if the vertex $LP$ is the nearest left vertex of a $Child$ vertex's parent, define $LP$ as the \textbf{Left Parent} of $Child$.

If $P$ is the Parent, $C$ is the Child, $LP$ is the Left Parent, $RP$ is the Right Parent, $ST$ is the subtree, $DST$ is the entire document structure tree, $RST$ is the right subtree, and $L$ is the section level. Use "$\leftarrow$" represents "assignment"; obviously, by analyzing the characteristics of the document structure tree, the following three basic properties can be summarized:

\noindent\textbf{PROPERTY 1}\quad In any $ST$ of $DST$, there is the following weight relationship:
\begin{equation}{
weight(P)<weight(C)<weight(P)
\label{eq2}
}
\end{equation}

\noindent\textbf{PROPERTY 2}\quad In any $ST$ of $DST$, there is the following weight relationship:
\begin{equation}
\begin{split}
&\forall\ vertex \in DST, \exists\ ST \subseteq DT, \\
&p \leftarrow Root(ST),rp \leftarrow Root(RST), \\
&\textbf{IF}\ weight(p)<weight(ve rtex)<weight(rp), \\
&\textbf{THEN}\ vertex \subseteq ST.
\end{split}
\label{eq3}
\end{equation}

\noindent\textbf{PROPERTY 3}\quad For any vertex in $DST$, its hierarchical attribution satisfies:
\begin{equation}
\begin{split}
&\forall\ vertex \in DST, \exists\ L_1,L_2,\ \textbf{and}\ L_2=L_1+1,\\
&\textbf{IF}\ \min\{weight(L_1)\}<weight(vertex)<\min\{weight(L_2)\},\\
&\textbf{THEN}\ vertex \subseteq L_1.
\end{split}
\label{eq4}
\end{equation}

The establishment order and traversal order of the document structure tree are consistent with the Chinese reading order, basically in the order that "root$\rightarrow$left node$\rightarrow$right". The establishment problem can be abstracted into the following representation:

\textbf{Known:} (1) the sections levels of vertices; (2) the weight of each vertex.

\textbf{Solve:} (1) the parent-child relationship of each vertex; (2) the hierarchy of vertices.

To make the computer able to organize metaknowledge elements, it is necessary to consider designing the data structure of the above-mentioned document structure tree model. The main problem of the design is to realize the relationship judgment of "parent$<$child$<$right parent" on the computer. To accomplish this task, it is needed to access each vertex from left to right and from top to bottom to judge the juxtaposition and inclusive relationship between left and right, superior and subordinate vertices (subtrees).

For the inequality "parent$<$child$<$right parent", the condition is incomplete. Because the parent node is traversed from top to bottom, the parent vertex must be accessed before the child vertex, that is, the left end of the inequality must be true. Therefore, it is only necessary to consider the case where the right end condition is incomplete, that is, the right parent vertex (right subtree) does not exist.

Obviously, if the classification discussion method is adopted, it is more expensive to add supplementary rules to where the right parent does not exist. Therefore, consider constructing the conditions that make the right part of the inequality always hold to adapt to the original rules, rather than establishing new rules. For this reason, the concept of "absolute right subtree" is introduced.

\vspace{0.3cm}
\noindent\textbf{DEFINITION 3}\quad The Absolute Right Subtree (ARS) is a document structure tree in which the root vertex weight is sufficiently large, and the child is empty. It is actually a fully weighted vertex at the rightmost end of the level. It only participates in weight comparisons but cannot be accessed.
\vspace{0.3cm}

The 4th layer belongs to the 4th section level, all are leaf vertices. Their subtrees are empty, so it is only necessary to establish an ARS in the 1st layer, the 2nd layer, and the 3rd layer. Moreover, by setting the traversal condition, the ARS can participate in the weight comparisons without being accessed, which solves the situation that the right parent vertex does not exist.

\begin{figure}[!h]
\centering
\includegraphics[width=\linewidth]{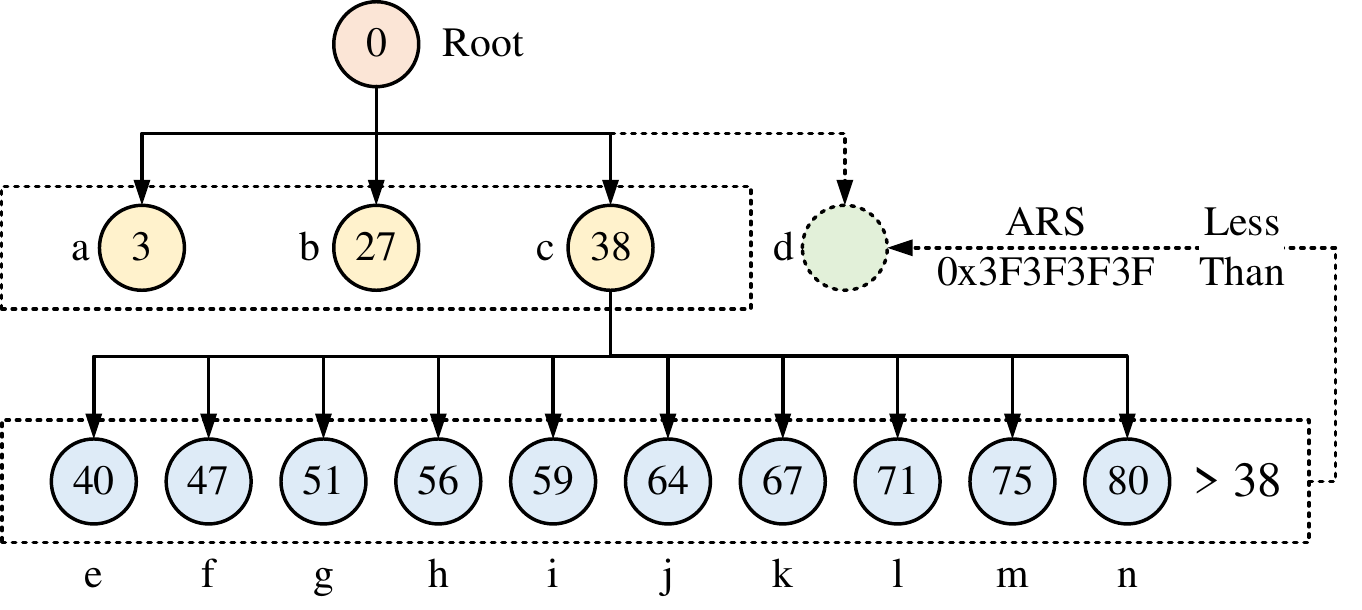}
\caption{Absolute Right Subtree: the vertices weights are part of the DST established by the extracted metaknowledge elements of the \emph{Report on the work of the Government, Year 2019, The Central People's Government, P. R. China}.}
\vspace{-0.4cm}
\label{fig_4}
\end{figure}

As is shown in Fig. 4, obviously, for all the child vertices $e$ to $n$ of vertex $c$, there is no right parent, and Eq. 2 is no longer valid. To ensure that Eq. 2 is always true, $weight(ARS)$ should be a sufficiently large number. In this work, 0x3F3F3F3F is set to the sufficiently large number, which not only avoids data overflow but also has the same order of magnitude as the maximum 0x7FFFFFFF of 32-bit integer data. Due to the introduction of the ARS $d$, the right parent of the child vertices $e$ to $n$ becomes $d$, and then the inequality $38<weight(subnodes)<0x3F3F3F3F$ holds.

Therefore, the document structure tree's minimum data unit is a structure containing the attributes of the root node and all the child vertices, which can be defined recursively to realize the document structure tree's construction.

\begin{lstlisting}[language={Python},numberstyle=\tiny,frame=single,basicstyle=\ttfamily]
class DST:
    def __init__(self):
        self.content = [] # Root
        self.subtree = [] # Child
\end{lstlisting}

\section{Experiments and Analysis}
To evaluate the metaknowledge elements extraction framework in this work, experiments are designed to verify the feasibility of extracting metaknowledge elements from multi-modal documents. Here we build governmental documents dataset GovDoc-CN to evaluate our framework. 

\subsection{Dataset and Setups}

In our previous work, we released a multi-modal governmental documents dataset GovDoc-CN\footnote{https://github.com/RuilinXu/GovDoc-CN}, which obtains 6 816 government document pages from the Central People's Government of China (CPGC) and its subordinate ministries and commissions from the policy document database of the CPGC.These documents are automatically typeset by \LaTeX \ and manually annotated in text modal and image modal.

Table 1 shows the statistical information of GovDoc-CN, in which there are 10 kinds of metaknowledge elements, including the sign of issuing authority, the document number, the title, the addressee, the 1st level section, the 2nd level section, the 3rd level section, the main text, the issuing authority and the date of writing. 

\begin{table}[h]
\caption{The Statistical Information Of GovDoc-CN}
\label{table1}
\begin{center}
\begin{tabular}{cc}
\toprule
Types  & \# Elements$^{\mathrm{a}}$   \\ \midrule
Sign of issuing authority & 1 347\\
Document   number                    & 1 344\\
Title                           & 1 359\\
Addressee                            & 1 065\\
1st level section                    & 5 184\\
2nd level section                    & 4 697\\
3rd level section                    & 1 415\\
Issuing authority                           & 2 073\\
Date of writing                  & 1 178\\
Paragraph                   & 10 280\\ 
\bottomrule
\multicolumn{2}{p{220pt}}{$^{\mathrm{a}}$The number of these elements comes from the text modal and image modal, and the annotations of the two modalities correspond to each other.}
\end{tabular}
\end{center}
\vspace{-0.5cm}
\end{table}

This work uses the same train, evaluation, and test set to train both the text modal model (BERTBASE+BiLSTM+CRF) and the image modal model (YOLOv4+PaddleOCR). Using 3 672 document pages for training, 918 pages for evaluation and 511 pages for testing. The setups are as follows:

\begin{itemize}
	\item[$\bullet$]\emph{Text Modal.} Using pre-trained Chinese BERT$_{\rm BASE}$ for fine-tuning. Batch size is set as 4, learning rate is 2e-5, max length is 512. Train 30 epochs. 
	\item[$\bullet$]\emph{Image Modal.} The YOLOv4 is trained with batch size 64, learning rate 1e-5, and is trained 20 000 iterations. 
\end{itemize}

\subsection{Metaknowledge Elements Extraction}

\begin{table}[!h]
\caption{Reuslts on GovDoc-CN}
\label{table3}
\begin{center}
\begin{tabular}{cccc}
\toprule
Metaknowledge Elements & \begin{tabular}[c]{@{}c@{}}BERT\\+BiLSTM\\+CRF\end{tabular} & \begin{tabular}[c]{@{}c@{}}YOLOv4\\ +PaddleOCR$^{\rm a}$\end{tabular} & MEF   \\\midrule
Sign of issuing authority                             & 0.9870                                                      & 0.8308                                                        & 0.9999 \\
Document number                                                  & 0.8977                                                      & 0.8588                                                        & 0.8977 \\
Title                                                            & 0.9290                                                      & 0.9211                                                        & 0.9299 \\
Addressee                                                        & 0.9608                                                      & 0.8090                                                        & 0.9905 \\
1st level section                                                & 0.9608                                                      & 0.7453                                                        & 0.8845 \\
2nd level section                                                & 0.7770                                                      & 0.6384                                                        & 0.8214 \\
3rd level section                                                & 0.7348                                                      & 0.6194                                                        & 0.8235 \\
Issuing authority                                                & 0.8646                                                      & 0.5342                                                        & 0.9347 \\
Date of writing                                                  & 0.8462                                                      & 0.8095                                                        & 0.9510 \\
Main text                                                        & 0.7097                                                      & 0.6452                                                        & 0.8146 \\
Average                                                          & 0.8668                                                      & 0.7412                                                        & 0.9048 \\\bottomrule
\multicolumn{4}{l}{$^{\rm a}$Considering OCR mismatch.}                                                                                                                                        
\end{tabular}
\end{center}
\end{table}

\begin{figure}[!h]
\centering
\includegraphics[width=\linewidth]{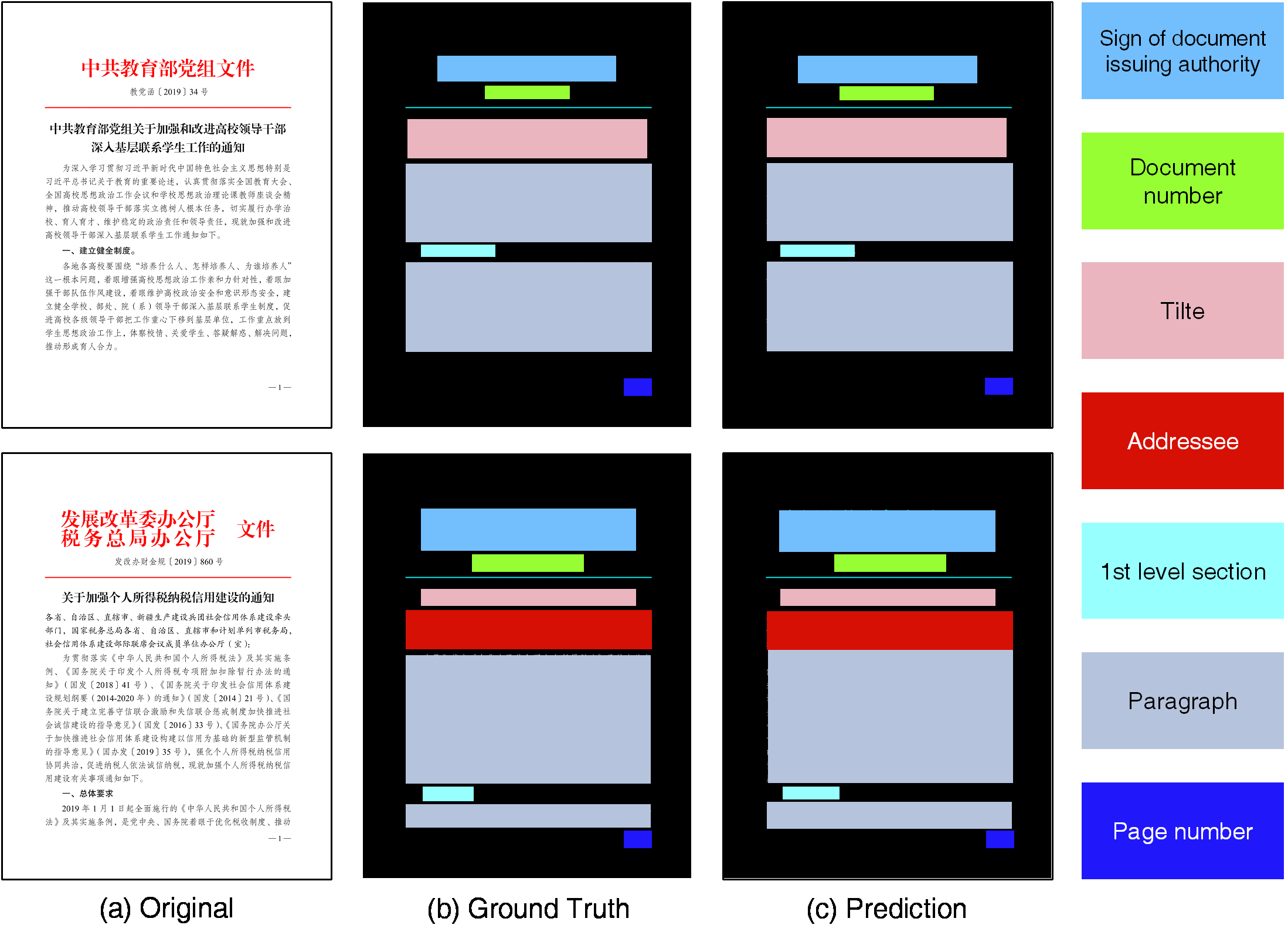}
\caption{Examples of original document pages, annotated ground truth pages, and outputs of the MEF.}
\vspace{-0.3cm}
\label{fig_5}
\end{figure}

The metaknowledge elements extraction results on GovDoc-CN has been shown in Table 2. Micro-F1 Score are used for evaluating our framework. Compared with the single-modal approaches BERT+BiLSTM+CRF and YOLOv4+PaddleOCR, our framework MEF reaches better performance in average F1-Score ( +3.80\% BERT+BiLSTM+CRF, +16.36\% YOLOv4+PaddleOCR ).  Table 2 also shows that the two single-modal approaches of text and image complement each other in extracting various metaknowledge elements. The example outputs are shown in Fig. 5.

As the layout structure of official documents is relatively simple, to verify the generalization ability of MEF to extract metaknowledge elements of complex documents, we originally planned to use large-scale multi-modal document layout analysis dataset DocBank \cite{docbank} for evaluation.

\begin{figure}[!b]
\centering
\includegraphics[width=\linewidth]{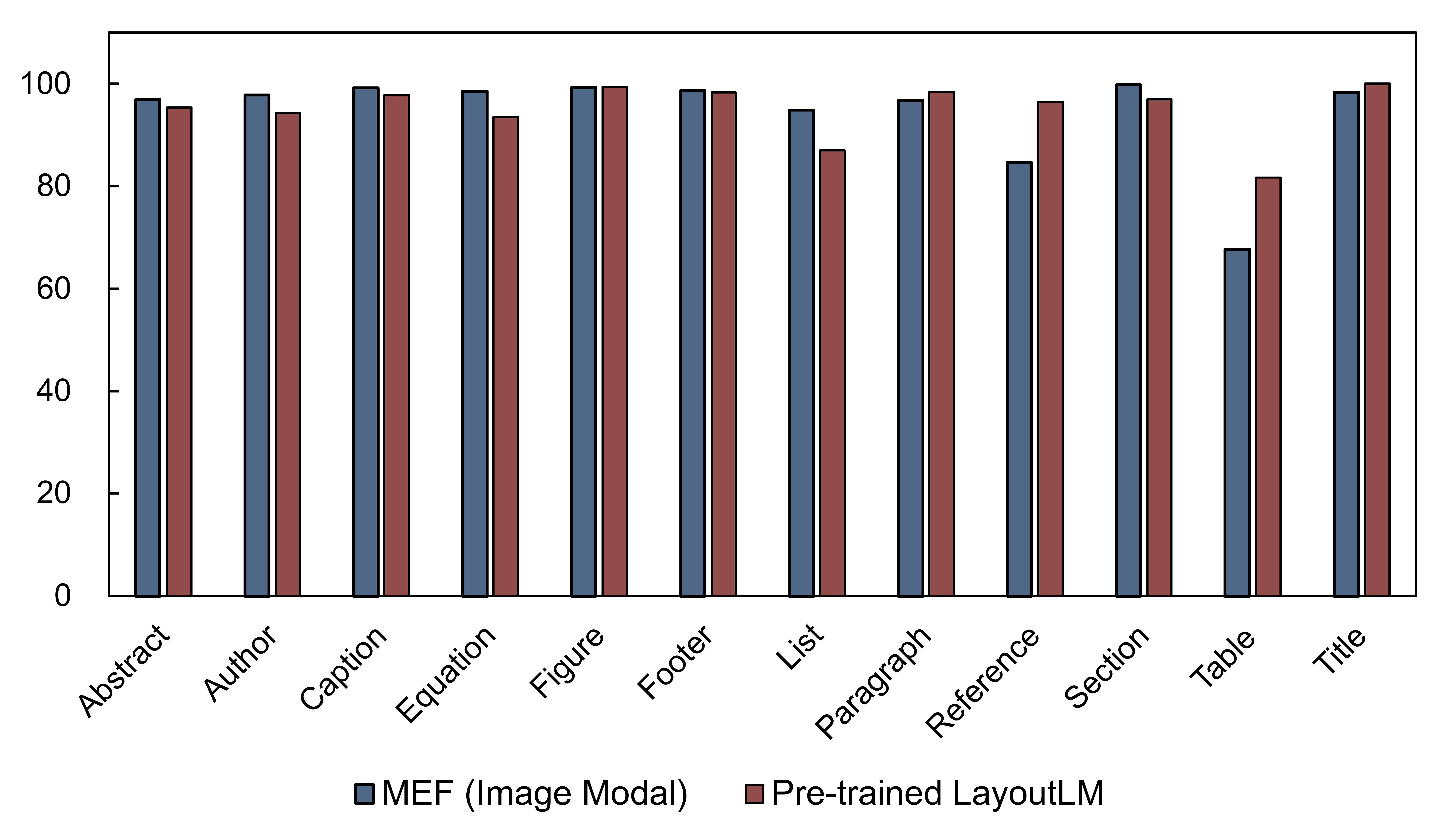}
\caption{The metaknowledge elements extraction results on DocBank\_50K.}
\vspace{-0.2cm}
\label{fig_6}
\end{figure}

However, when using the dataset, we find that the computer annotating approach has a terrible impact on the data quality, and there is a lot of dirty data mixed in the bounding boxes of the image modal annotations. In the text modal annotations, the dataset only gives the label of each token but does not provide the starting and ending position of paragraphs, making it very difficult to convert these annotated tokens into BIO format data that commonly used in NLP. Limited by the computing capability, we have to run our experiments on a subset of 50 000 samples of DocBank. We find that the the text modal results are deplorable according to either the text classification or the NER approaches. In the image modal, after a large amount of complex bounding boxes merging work that has transformed the token-level bounding boxes into the component-level, we finally get SOTA (Table 3). 

As shown in Table 3 and Fig. 6, MEF's extraction performance of 7 types of elements exceeds that of LayoutLM trained with 10 times more samples. Due to issues of the DocBank, MEF's performance in the text modal is not as good as pleasant. Nevertheless, considering its performance on GovDoc-CN, we believe that switching to a better multi-modal document dataset will achieve more ideal results. Unfortunately, there is currently a lack of relevant datasets. The layout analysis datasets represented by PubLayNet \cite{publaynet} have only a single modal of image. Therefore, we will work harder to strengthen the weak links in the next work and expand GovDoc-CN into a multilingual and multi-modal document dataset with more document types.

\begin{table}[!h]
\caption{Reuslts on DocBank\_50K}
\label{table3}
\begin{center}
\setlength{\tabcolsep}{3mm}
\begin{tabular}{ccc}
\toprule
\begin{tabular}[c]{@{}c@{}}Metaknowledge\\ Elements\end{tabular} & \begin{tabular}[c]{@{}c@{}}MEF\\ (Image Modal)\end{tabular} & \begin{tabular}[c]{@{}c@{}}Pre-trained\\ LayoutLM$^{\rm a}$\end{tabular} \\ \midrule
Abstract                                                         & \textbf{0.9693}                                              & 0.9537                                                         \\
Author                                                           & \textbf{0.9778}                                              & 0.9423                                                         \\
Caption                                                          & \textbf{0.9912}                                              & 0.9788                                                         \\
Equation                                                         & \textbf{0.9858}                                              & 0.9346                                                         \\
Figure                                                           & 0.9934                                                       & \textbf{0.9941}                                                \\
Footer                                                           & \textbf{0.9874}                                              & 0.9826                                                         \\
List                                                             & \textbf{0.9481}                                              & 0.8699                                                         \\
Paragraph                                                        & 0.9665                                                       & \textbf{0.9849}                                                \\
Reference                                                        & 0.8461                                                       & 0.9643                                                         \\
Section                                                          & \textbf{0.9977}                                              & 0.9694                                                         \\
Table                                                            & 0.6764                                                       & \textbf{0.8175}                                                \\
Title                                                            & 0.9834                                                       & \textbf{0.9999}           \\ \bottomrule       
\multicolumn{3}{p{220pt}}{$^{\rm a}$ Results come from Ref. \cite{docbank}, the model was trained and evaluated on the complete DocBank.}
\end{tabular}
\end{center}
\vspace{-0.2cm}
\end{table}

\subsection{Generating Metaknowldge}
The previous processes have acquired the hierarchical structure of documents, that is, the typologies of documents metaknowledge. However, the semantic information has not been linked to document metaknowledge yet. Therefore, this work picks Ref. \cite{yolo1} from DocBank\_50K as an example to demonstrate the whole process of how to generate metaknowledge from documents.

The metaknowledge elements are extracted through MEF and organized by the DST model. Furthermore, entity-relation triples, or the \textit{Structure Contextual Triples} called in Ref. \cite{contexttriples}, are extracted from the paragraphs and linked to the sections or subsections they belong to, and then we get the metaknowledge of Ref. \cite{yolo1}. The metaknowledge is saved in the Neo4j database, which is shown in Fig. 7(a).

\begin{figure}[htbp]
\centering
\subfloat[{Metaknowledge of Ref. \cite{yolo1} which is displayed in Neo4j}]{\label{fig:a}\includegraphics[width=\linewidth]{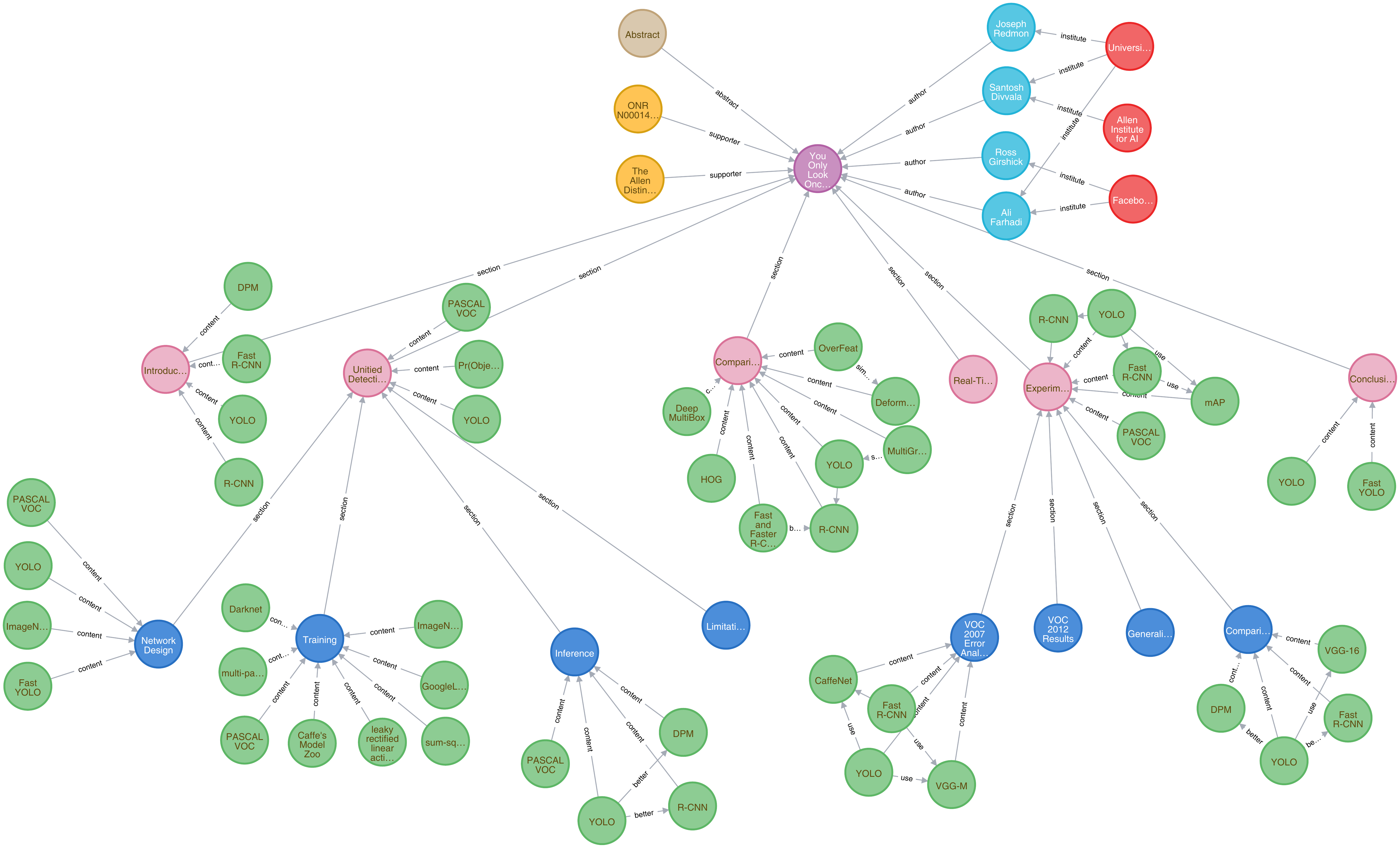}}\qquad\\
\subfloat[{The hierarchical structure of Ref. \cite{yolo1}'s metaknowledge}]{\label{fig:b}\includegraphics[width=\linewidth]{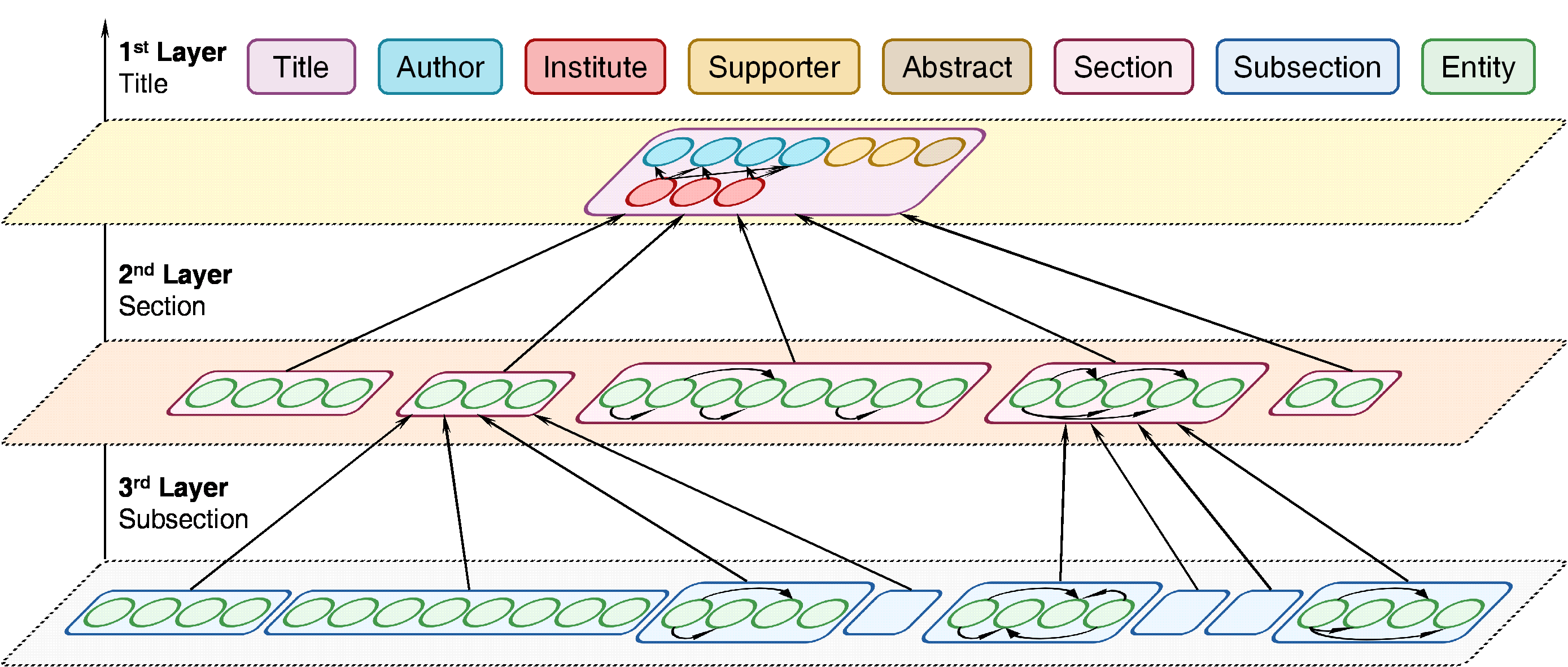}}\\	
\caption{An example: metaknowledge of Ref. \cite{yolo1}}
\vspace{-0.4cm}
\label{fig_7}
\end{figure}

We explain the metaknowledge of Ref. \cite{yolo1} in Fig. 7(b). Metaknowledge is a graph structure with hierarchical characteristics, which is basically consistent with the directory structure of documents. In Fig. 7(b), the hierarchical characteristics are represented by three layers: the 1st layer, the title; the 2nd layer, the sections; and the 3rd layer, the subsections. In the 1st layer, metaknowledge elements describe the attributes of documents such as the authors, the institutes, and the supporters. These metaknowledge elements are represented as a form similar to triple-based knowledge. In the 2nd and 3rd layers, we extract entities and relations from paragraphs, which consists of triple-based knowledge, and then linked to their nearest section or subsection they belong to.

This work recognizes that the citation network is also well-organized in metaknowledge. As shown in Fig. 8, all the references of Ref. \cite{yolo1} are extracted and organized by hierarchical structure. Compared with the current citation network analysis research, the citation network based on metaknowledge no longer takes the authors as the main research target, but pays more attention to the knowledge structure and knowledge itself. Each reference is a document that can generate metaknowledge. Therefore, a large-scale metaknowledge network can be constructed by converting a large number of references into metaknowledge.

\begin{figure}[!h]
\centering
\includegraphics[width=0.8\linewidth]{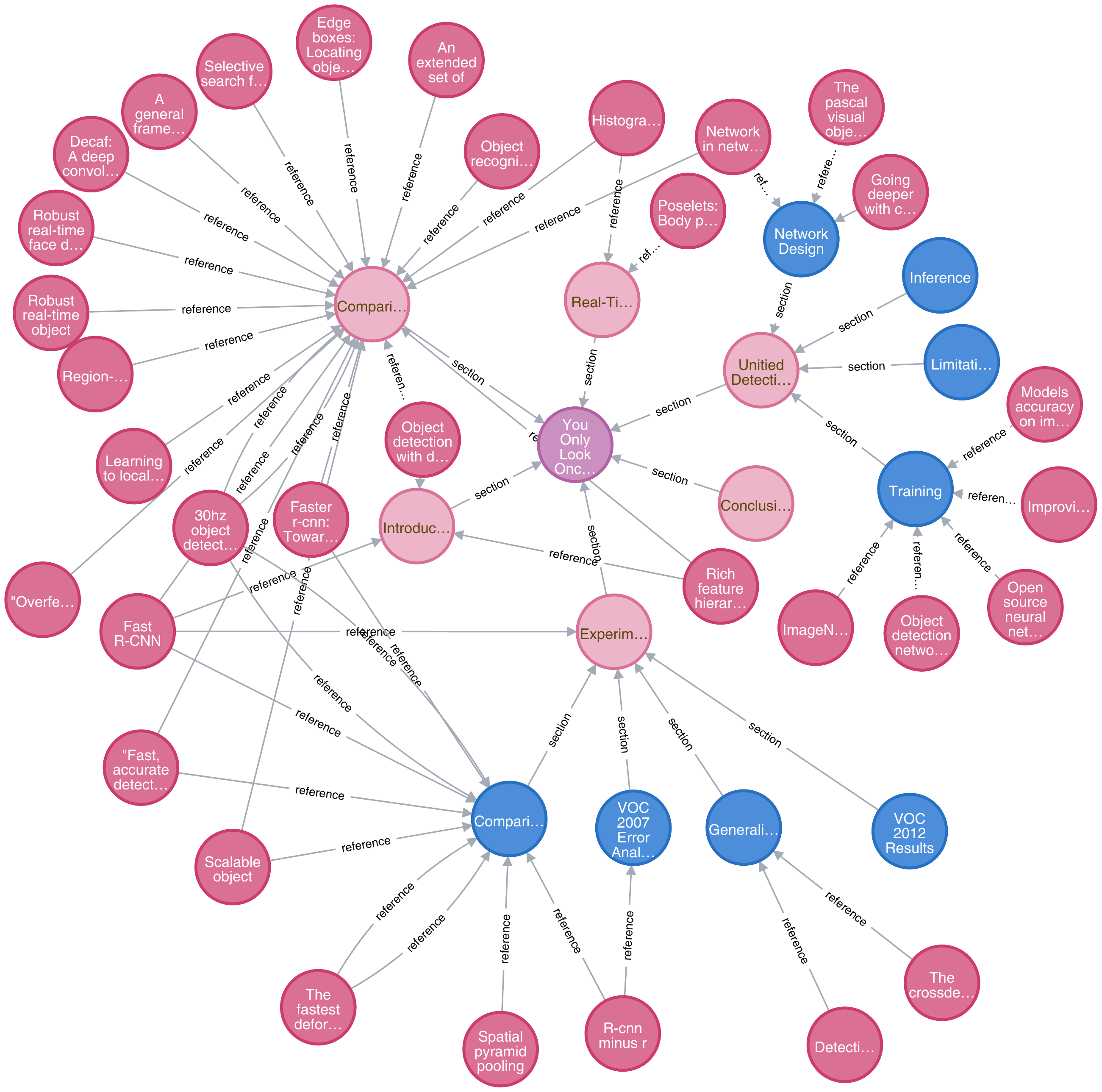}
\caption{The reference network of Ref. \cite{yolo1} constructed by metaknowledge.}
\vspace{-0.4cm}
\label{fig_8}
\end{figure}

\begin{figure*}[!h]
\centering
\includegraphics[width=\linewidth]{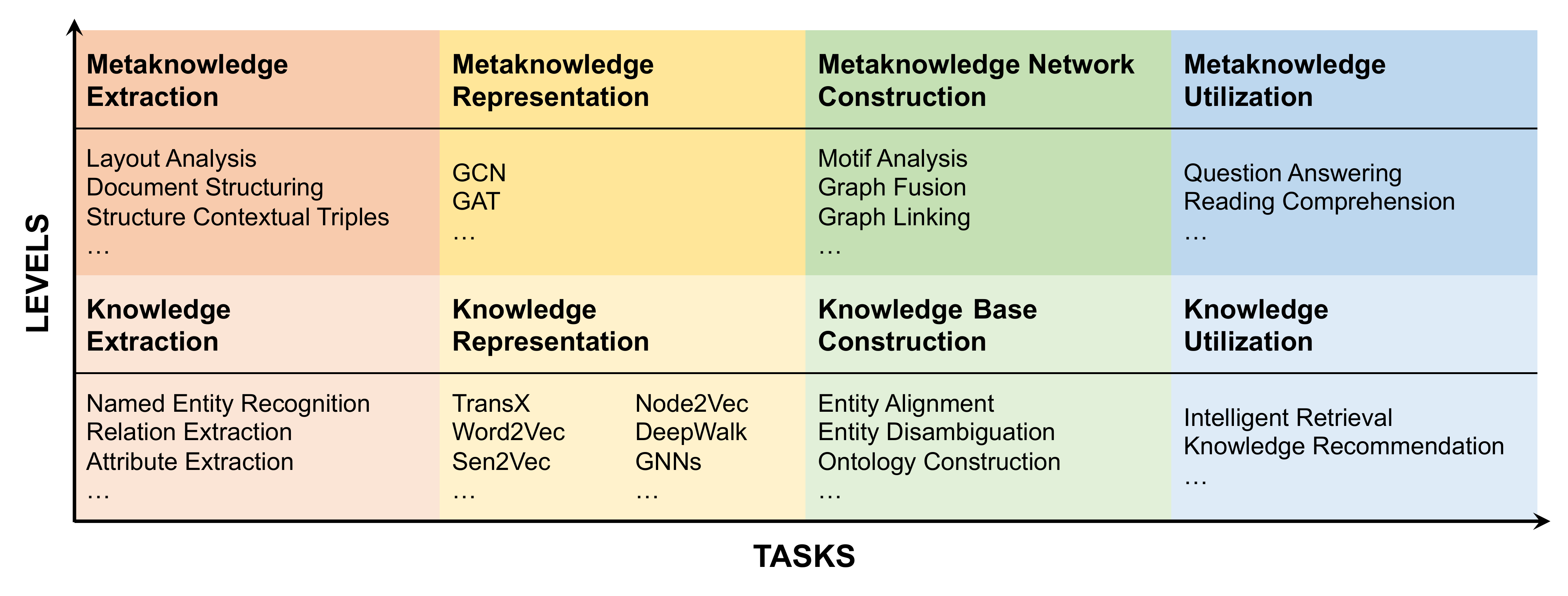}
\caption{The task flow of metaknowledge applications. The horizontal axis represents the task flow of knowledge and metaknowledge, and the vertical axis represents the low-to-high levels of knowledge and metaknowledge.}
\vspace{-0.2cm}
\label{fig_9}
\end{figure*}

\section{Task Flow Of Metaknowledge Applications}
Metaknowledge is the knowledge about knowledge, in other words, metaknowledge realizes the organization and management of specific knowledge. Under the metaknowledge empowerment, the knowledge can be effectively used at a higher level. We analyze the whole task flow of current knowledge engineering applications, inspired by that, we propose a novel task flow of metaknowledge applications (Fig. 9).

Fig. 9 shows the task flow of knowledge and metaknowledge from two different levels. From the perspective of tasks, the knowledge application task flow can be divided into four parts: (1) Knowledge Extraction, including named entity recognition, relation extraction, attribute extraction, etc. (2) Knowledge Representation. Generally, it has two types: one is based on entity-relation triples, where the approaches include TransX \cite{transe,transd,transh}, Word2Vec \cite{word2vec}, Sen2Vec \cite{sen2vec}, Para2Vec \cite{para2vec}; the other type is based on knowledge networks, where the approaches include Node2Vec \cite{node2vec}, DeepWalk \cite{deepwalk}, Graph Neural Networks (GNNs) \cite{gnns}, etc. (3) Knowledge Base Construction, including entity alignment, entity disambiguation, ontology construction, knowledge reasoning, etc. (4) Knowledge Utilization, including intelligent retrieval, knowledge recommendation, etc. 

The metaknowledge application task flow can also be divided into four parts: (1) Metaknowledge Extraction, the data generally comes from multiple modes, but the approaches basically include document layout analysis, document structuring, structure contextual triples extraction, etc. (2) Metaknowledge Representation. Considering metaknowledge is weighted and directional graph data, theoretically the approaches such as graph convolutional network (GCN) \cite{gcn1,gcn2} and graph attention network (GAT) \cite{gat} could be well-performed on this task. (3) Metaknowledge Networks Construction. Its main task is building a network from large amounts of metaknowledge, using approaches such as graph fusion, graph linking, etc. (4) Metaknowledge Utilization, which includes Metaknowledge-based Question Answering (MbQA) and Metaknowledge-based Document Understanding (MbDU), among which MbQA seems to be a combination of knowledge base question answering (KBQA) \cite{kbqa1,kbqa2} and document-based question answering (DBQA) \cite{dbqa1,dbqa2}, it considers both semantic logic of the knowledge and hierarchical structure of the document. 

From the perspective of levels (Fig. 9), metaknowledge applications is higher, more holistic and more macroscopic than knowledge applications. The information elements of knowledge are entities and the relations (entity-relation triples), while the information elements of metaknowledge should be knowledge and the relations between knowledge (structure contextual triples). Specifically, metaknowledge is the weighted and directional graph data, which could have a better performance than the triple-based knowledge bases on tasks such as question answering, machine reading comprehension, etc. Knowledge applications realize the management of information, similarly, metaknowledge applications will realize the management and organization of knowledge. 



\section{Conclusion}
In this work, we introduce the concept of metaknowledge to knowledge engineering researches, propose a framework MEF to extract metaknowledge elements from multi-modal documents, and the Document Structure Tree model to organize metaknowledge with document topological and hierarchical features. Experiments and analysis demonstrate the effectiveness of MEF, an metaknowledge example of Ref. \cite{yolo1} is given to analyze what metaknowledge actually is and how is it organized. Finally, our work proposes the task flow of metaknowledge applications, analyzes the tasks from upstream to downstream that metaknowledge is utilized. Future work will focus on metaknowledge representation learning, metaknowledge network construction, and how to verify the functionality of metaknowledge. 

\bibliographystyle{IEEEtran}
\bibliography{references}{}

\end{document}